\documentstyle{article}

\makeatletter
  \newlength\titlebox 
 \twocolumn 

\def\addcontentsline#1#2#3{}

\def\maketitle{\par
 \begingroup
   \def\thefootnote{\fnsymbol{footnote}}
   \def\@makefnmark{\hbox to 0pt{$^{\@thefnmark}$\hss}}
   \twocolumn[\@maketitle] \@thanks
 \endgroup
 \setcounter{footnote}{0}
 \let\maketitle\relax \let\@maketitle\relax
 \gdef\@thanks{}\gdef\@author{}\gdef\@title{}\let\thanks\relax}
\def\@maketitle{\vbox to \titlebox{\hsize\textwidth
 \linewidth\hsize \vskip 0.625in minus 0.125in \centering
 {\huge\bf \@title \par} \vskip 0.2in plus 1fil minus 0.1in
 {\def\and{\unskip\enspace{\rm and}\enspace}%
  \def\And{\end{tabular}\hss \egroup \hskip 1in plus 2fil 
           \hbox to 0pt\bgroup\hss \begin{tabular}[t]{c}\Large\bf}%
  \def\AND{\end{tabular}\hss\egroup \hfil\hfil\egroup
	  \vskip 0.25in plus 1fil minus 0.125in
	   \hbox to \linewidth\bgroup\Large \hfil\hfil
 	     \hbox to 0pt\bgroup\hss \begin{tabular}[t]{c}\Large\bf}
  \hbox to \linewidth\bgroup\Large \hfil\hfil
    \hbox to 0pt\bgroup\hss \begin{tabular}[t]{c}\Large\bf\@author 
			    \end{tabular}\hss\egroup
    \hfil\hfil\egroup}
  \vskip 0.3in plus 2fil minus 0.1in
}}

\def\section{\@startsection {section}{1}{\z@}{-2.0ex plus
    -0.5ex minus -.2ex}{3pt plus 2pt minus 1pt}{\Large\bf\centering}}
\def\subsection{\@startsection{subsection}{2}{\z@}{-2.0ex plus
    -0.5ex minus -.2ex}{3pt plus 2pt minus 1pt}{\large\bf\raggedright}}
\def\subsubsection{\@startsection{subparagraph}{3}{\z@}{-6pt plus
   2pt minus 1pt}{-1em}{\normalsize\bf}}

\skip\footins 9pt plus 4pt minus 2pt
\def\footnoterule{\kern-3pt \hrule width 5pc \kern 2.6pt }
\labelwidth\leftmargini\advance\labelwidth-\labelsep \labelsep 5pt

\def\@listi{\leftmargin\leftmargini}
\def\@listii{\leftmargin\leftmarginii
   \labelwidth\leftmarginii\advance\labelwidth-\labelsep
   \topsep 2pt plus 1pt minus 0.5pt
   \parsep 1pt plus 0.5pt minus 0.5pt
   \itemsep \parsep}
\def\@listiii{\leftmargin\leftmarginiii
    \labelwidth\leftmarginiii\advance\labelwidth-\labelsep
    \topsep 1pt plus 0.5pt minus 0.5pt 
    \parsep \z@ \partopsep 0.5pt plus 0pt minus 0.5pt
    \itemsep \topsep}
\def\@listiv{\leftmargin\leftmarginiv
     \labelwidth\leftmarginiv\advance\labelwidth-\labelsep}
\def\@listv{\leftmargin\leftmarginv
     \labelwidth\leftmarginv\advance\labelwidth-\labelsep}
\def\@listvi{\leftmargin\leftmarginvi
     \labelwidth\leftmarginvi\advance\labelwidth-\labelsep}

\belowdisplayskip \abovedisplayskip
\def\@normalsize{\@setsize\normalsize{11pt}\xpt\@xpt}
\def\small{\@setsize\small{10pt}\ixpt\@ixpt}
\def\footnotesize{\@setsize\footnotesize{10pt}\ixpt\@ixpt}
\def\scriptsize{\@setsize\scriptsize{8pt}\viipt\@viipt}
\def\tiny{\@setsize\tiny{7pt}\vipt\@vipt}
\def\large{\@setsize\large{12pt}\xipt\@xipt}
\def\Large{\@setsize\Large{14pt}\xiipt\@xiipt}
\def\LARGE{\@setsize\LARGE{16pt}\xivpt\@xivpt}
\def\huge{\@setsize\huge{20pt}\xviipt\@xviipt}
\def\Huge{\@setsize\Huge{23pt}\xxpt\@xxpt}

\def\@citex[#1]#2{\if@filesw\immediate\write\@auxout{\string\citation{#2}}\fi
  \def\@citea{}\@cite{\@for\@citeb:=#2\do
    {\@citea\def\@citea{;\penalty\@m\ }\@ifundefined
       {b@\@citeb}{{\bf ?}\@warning
       {Citation `\@citeb' on page \thepage \space undefined}}%
{\csname b@\@citeb\endcsname}}}{#1}}
\let\@internalcite\cite
\def\cite{\def\citename##1{##1, }\@internalcite}
\def\shortcite{\def\citename##1{}\@internalcite}
\def\newcite{\leavevmode\def\citename##1{{##1} (}\@internalciteb}

\def\@citexb[#1]#2{\if@filesw\immediate\write\@auxout{\string\citation{#2}}\fi
  \def\@citea{}\@newcite{\@for\@citeb:=#2\do
    {\@citea\def\@citea{;\penalty\@m\ }\@ifundefined
       {b@\@citeb}{{\bf ?}\@warning
       {Citation `\@citeb' on page \thepage \space undefined}}%
\hbox{\csname b@\@citeb\endcsname}}}{#1}}
\def\@internalciteb{\@ifnextchar [{\@tempswatrue\@citexb}{\@tempswafalse\@citexb[]}}

\def\@newcite#1#2{{#1\if@tempswa, #2\fi)}}

\def\@biblabel#1{\def\citename##1{##1}[#1]\hfill}

\def\@cite#1#2{({#1\if@tempswa , #2\fi})}

\def\thebibliography#1{\vskip\parskip%
\vskip\baselineskip%
\def\baselinestretch{1}%
\ifx\@currsize\normalsize\@normalsize\else\@currsize\fi%
\vskip-\parskip%
\vskip-\baselineskip%
\section*{References\@mkboth
 {References}{References}}\list
 {}{\setlength{\labelwidth}{0pt}\setlength{\leftmargin}{\parindent}
 \setlength{\itemindent}{-\parindent}}
 \def\newblock{\hskip .11em plus .33em minus -.07em}
 \sloppy\clubpenalty4000\widowpenalty4000
 \sfcode`\.=1000\relax}

\def\@lbibitem[#1]#2{\item[]\if@filesw 
      { \def\protect##1{\string ##1\space}\immediate
        \write\@auxout{\string\bibcite{#2}{#1}}\fi\ignorespaces}}

\def\@bibitem#1{\item\if@filesw \immediate\write\@auxout
       {\string\bibcite{#1}{\the\c@enumi}}\fi\ignorespaces}
\makeatother

\newcommand{\f}[1]{\mathord{\mbox{\rm #1}}}
\newcommand{\tup}[1]%
  {\left\langle
#1\right\rangle}
\newtheorem{lemma}{Lemma}
\newenvironment{proofof}[1]%
  {\nobreak \trivlist \item[\hskip \labelsep{\bf Proof{\rm\ (#1)}:}]}%
  {\nobreak\hfill\nobreak$\Box$\\*}
\newcommand{\set}[1]{\left\{
#1\right\}}
\newcommand{\pair}[2]%
  {\left\langle
#1,#2\right\rangle}
\newcommand{\sP}{{\cal P}}
\newcommand{\power}{{\bf \sP}}
\def\defiff{\mathrel{\buildrel \rm def \over \Longleftrightarrow}}
\newcommand{\tor}{\mbox{ or }}
\newcommand{\tand}{\mbox{ and }}
\newcommand{\sA}{{\cal A}}
\newtheorem{definition}{Definition}
\newtheorem{proposition}{Proposition}
\newenvironment{proof}%
  {\nobreak \trivlist \item[\hskip \labelsep{\bf Proof:}]}%
  {\nobreak\hfill\nobreak$\Box$\\*}
\newtheorem{corollary}{Corollary}

\begin{document}

\bibliographystyle{acl}

\author{James Rogers\thanks{The work reported here
owes a great deal to extensive discussions with K. Vijay-Shanker.}\\
Dept. of Computer and Information Sciences\\
University of Delaware\\
Newark, DE 19716, USA\\
{\tt jrogers@cis.udel.edu}}

\title{Capturing CFLs with Tree Adjoining Grammars}

\maketitle

\begin{abstract}
We define a decidable class of TAGs that is strongly equivalent
to CFGs and is cubic-time parsable. This class serves to lexicalize CFGs
in the same manner as the LCFGs of Schabes and Waters but with
considerably less  restriction on the form of the grammars.  The class
provides a normal form for TAGs that generate local sets in much the
same way that regular grammars provide a normal form for CFGs that
generate regular sets.
\end{abstract}

\section{Introduction}\label{sec.intro}
We introduce the notion of {\em Regular Form} for Tree Adjoining
Grammars (TAGs).  The class of TAGs that are in regular from
is equivalent in strong generative capacity\footnote{We will
refer to equivalence of
the sets of trees generated by two grammars or classes of grammars as
{\em strong equivalence}.  Equivalence of their string languages will be
referred to as {\em weak equivalence}.}
to the Context-Free
Grammars, that is, the sets of trees generated
by TAGs in this class are the {\em local sets}---the 
sets of derivation trees generated by CFGs.\footnote{Technically, the
sets of
trees generated by TAGs in the class are {\em recognizable sets}.  The
local and recognizable sets are equivalent modulo projection.  We
discuss the
distinction in the next section.}
Our investigations were initially
motivated by the work of Schabes, Joshi, and Waters in lexicalization of
CFGs via TAGs~\cite{SchJos91,JosSch92,SchWat93a,SchWat93b,schabe90}.
The class we describe not only serves to lexicalize CFGs
in
a way that is more faithful and more flexible in its encoding
than earlier work,
but provides a basis for using
the more expressive TAG formalism to define Context-Free Languages (CFLs.)

In
\newcite{ScAbJo88} and 
\newcite{schabe90} a general notion of {\em lexicalized
grammars} is introduced.  A grammar is lexicalized in this sense if each
of the basic structures it manipulates is associated with a lexical item,
its {\em anchor}.  The set of structures relevant to a particular input
string, then, is selected by the lexical items that occur in that
string.  There are a number of reasons for exploring lexicalized
grammars.  Chief among these are linguistic considerations---lexicalized
grammars reflect the tendency in many current syntactic theories to
have the details of the syntactic structure be projected from the
lexicon.  There are also practical advantages.  All lexicalized grammars
are finitely ambiguous and, consequently, recognition for them is
decidable.  Further, lexicalization supports strategies that can, in
practice, improve the speed of recognition algorithms~\cite{ScAbJo88}.

One grammar formalism
is said to {\em lexicalize} another~\cite{JosSch92} if for every grammar
in the second formalism there is a lexicalized grammar in the first
that generates exactly the same set of structures.  While CFGs are
attractive for efficiency of recognition,
\newcite{JosSch92} 
have shown that an arbitrary CFG cannot, in general,
be converted into a strongly equivalent lexicalized CFG. 
Instead, they show how CFGs can be lexicalized by LTAGS
(Lexicalized TAGs).  While the LTAG that lexicalizes a given CFG must be
strongly equivalent to that CFG, both
the languages and sets of trees generated by LTAGs as a class
are strict supersets of the CFLs and local sets.  Thus, while
this gives a means of constructing a lexicalized grammar
from an existing CFG,  it does not provide a direct method for
constructing lexicalized grammars that are known to be equivalent to 
(unspecified) CFGs.  Furthermore, the best known
recognition algorithm for LTAGs runs in $\f{O}(n^6)$ time.

Schabes and Waters~\shortcite{SchWat93a,SchWat93b}  define Lexicalized
Context-Free Grammars  (LCFGs), a class of lexicalized TAGs (with 
restricted adjunction) that not only lexicalizes CFGs,
but is cubic-time parsable
and is {\em weakly} equivalent to CFGs.  These LCFGs have a couple of
shortcomings.  First, they are not strongly equivalent to CFGs.  Since
they are cubic-time parsable this is primarily a
theoretical rather than practical
concern.  More importantly, they employ structures of a highly
restricted form.  Thus the restrictions of the formalism, in some
cases, may override linguistic considerations in constructing the
grammar.  Clearly any class of TAGs that are cubic-time parsable, or that
are equivalent in any sense to CFGs, must be restricted in some way. 
The question is what restrictions are necessary.

In this paper we directly address the issue of identifying a class of
TAGs that are strongly equivalent to CFGs.  In doing so we define such
a class---TAGs in {\em regular form}---that is decidable, cubic-time
parsable, and lexicalizes CFGs.  Further, regular form is essentially a
closure condition on the elementary trees of the TAG.  Rather than
restricting the form of the trees that can be employed, or the
mechanisms by which they are combined, it requires that whenever a
tree with a particular form can be derived then certain other
related trees must be
derivable as well.  The algorithm for deciding whether a given grammar
is in regular form can produce a set of elementary trees that will
extend a grammar that does not meet the condition to one that 
does.\footnote{Although the result of this process is not, in general,
equivalent to the original grammar.}
Thus the grammar can be written largely on the basis of the linguistic
structures that it is intended to capture.  We show that, while the
LCFGs that are built by Schabes and Waters's algorithm for
lexicalization of CFGs are in regular form, the restrictions they employ
are unnecessarily strong.

Regular form provides a partial answer to the more general issue of
characterizing the TAGs that generate local sets.  It serves as a
normal form for these TAGs in the same way that regular grammars serve as a
normal form for CFGs that generate regular languages.
While for every TAG that generates a local set there is a
TAG in regular form that generates the same set, and every TAG in
regular form generates a local set (modulo projection),  there
are TAGs that are not in regular form that generate local
sets, just as there are CFGs that generate regular languages that are
not regular grammars.

The next section of this paper briefly introduces notation for TAGs and
the concept of recognizable sets.  Our results on regular form are
developed in the subsequent section.  We first define a restricted use
of the adjunction operation---derivation by {\em regular
adjunction}---which we show derives only recognizable sets. We then
define the class of TAGs in regular form and show that the set of trees
derivable in a TAG of this form is derivable by regular adjunction in
that TAG and is therefore recognizable.  We next show that every local
set can be generated by a TAG in regular form and that Schabes and
Waters's construction for LCFGs in fact produces TAGs in regular form. 
Finally, we provide an algorithm for deciding if a given TAG is in
regular form.  We close with a discussion of  the implications of this
work with respect to the lexicalization of CFGs and the use of TAGs
to define languages that are strictly context-free, and raise the
question of whether our results can be strengthened for some classes of
TAGs.

\section{Preliminaries}\label{sec.foundations}
\subsection{Tree Adjoining Grammars}\label{sec.tags}
Formally, a TAG is a five-tuple $\tup{\Sigma,\f{NT},I,A,S}$ where:
\\[1ex]
\begin{tabular}{rl}
$\Sigma$ & is a finite set of
{\em terminal symbols},\\
$\f{NT}$ & is a finite set of
{\em non-terminal symbols},\\
{$I$} & is a finite set of
{\em elementary initial trees},\\
{$A$} & is a finite set of
{\em elementary auxiliary trees},\\
{$S$} & is a distinguished non-terminal,\\
&\quad the {\em start symbol}.
\end{tabular}
\vspace{1ex}

Every non-frontier node of a tree in $I\cup A$ is labeled with a
non-terminal.
Frontier nodes may be labeled with either a terminal or a
non-terminal.
Every tree in $A$ has exactly one frontier node that is
designated as its {\em foot}.  This must be labeled with the
same non-terminal as the root.  The auxiliary and initial trees are
distinguished by the presence (or absence, respectively) of a foot node. 
Every other frontier node that is labeled with a non-terminal is
considered to be {\em marked for substitution.}
In a lexicalized TAG (LTAG) every tree in $I\cup A$ must have some
frontier node designated the {\em anchor}, which must be
labeled with a terminal.

Unless otherwise stated, we
include both elementary and derived trees when referring to initial
trees and auxiliary trees.
A TAG derives trees by a sequence of substitutions and adjunctions in
the elementary trees.  In {\em substitution} an instance of an
{\em initial tree} in which the root is labeled $X\in\f{NT}$ is
substituted for a frontier node (other than the foot) in an instance of
either an initial or auxiliary tree that is also
labeled $X$.  Both trees may be either
an elementary tree or a derived tree.

In {\em adjunction} an instance of an {\em auxiliary tree}
in which the root and foot are labeled $X$ is inserted at
a node,
also labeled $X$, 
in an instance of either an initial or auxiliary tree as follows:
the subtree at that node is excised, the auxiliary
tree is substituted at that node, and the excised subtree is substituted
at the foot of the auxiliary tree.  Again, the trees may be either
elementary or derived.

The set of objects ultimately derived by a TAG $G$ is $T(G)$, the set of
{\em completed} initial 
trees derivable in $G$.  These are the initial trees derivable in $G$ in
which the root is labeled $S$ and
every frontier node is labeled with a terminal
(thus no nodes are marked for substitution.) 
We refer to the set of all trees, both initial and
auxiliary, with or without nodes marked for substitution,
that are
derivable in $G$ as $T'(G)$.  The {\em language} derived by $G$ is $L(G)$
the
set of strings in $\Sigma^*$ that are the yields of trees in $T(G)$.

In this paper, all TAGs are {\em pure} TAGs, i.e., without
adjoining constraints.  Most of our results go through for TAGs with
adjoining constraints as well, but there is much more to say about these
TAGs and the implications of this work in distinguishing the pure TAGs
from TAGs in general.  This is a part of our ongoing research.

The path between the root and foot (inclusive)
of an auxiliary tree is referred to as
its {\em spine}.
Auxiliary trees in which no node on
the spine other than the foot is labeled with the same non-terminal as the
root we call a {\em proper} auxiliary tree.

\begin{lemma}\label{lem.proper}
For any TAG $G$ there is a TAG $G'$ that includes no improper
elementary trees such that
$T(G)$ is a projection of $T(G')$.
\end{lemma}
\begin{proofof}{Sketch}
The grammar $G$ can be relabeled with
symbols in $\set{\pair{x}{i}\mid x\in\Sigma\cup\f{NT},i\in\set{0,1}}$
to form $G'$.
Every auxiliary tree is duplicated, with the root and foot labeled
$\pair{X}{0}$ in one copy and $\pair{X}{1}$ in the other.
Improper elementary auxiliary trees can be avoided by appropriate choice
of labels along the spine.
\end{proofof}

The labels in
the trees generated by $G'$ are a refinement of the labels
of the trees generated by $G$.  Thus $G'$ partitions the categories
assigned by $G$ into subcategories on the basis of (a fixed amount of)
context.  While the use here is technical rather than natural,
the approach is familiar, as in the use of slashed
categories to handle movement.

\subsection{Recognizable Sets}\label{sec.automata}
The local sets are formally very closely related to the recognizable
sets, which are somewhat more convenient to work with.
These are sets of trees that are accepted by {\em finite-state tree
automata}~\cite{GecSte84}.  
If $\Sigma$ is a finite alphabet, a {\em $\Sigma$-valued tree} is a
finite, rooted, left-to-right ordered
tree, the nodes of which are labeled with symbols in $\Sigma$.  We will
denote such a tree in which the root is labeled $\sigma$ and in which
the subtrees at the children of the root are $t_1,\ldots,t_n$ as
$\sigma(t_1,\ldots,t_n)$. The set
of all $\Sigma$-valued trees is denoted $T_\Sigma$.

A {\em (non-deterministic)
bottom-up finite state tree automaton} over $\Sigma$-valued trees
is a tuple $\tup{\Sigma, Q, M, F}$ where:
\\[1ex]
\begin{tabular}{rp{2.75in}}
{$\Sigma$} & is a finite {\em alphabet},\\
{$Q$} & is a finite set of {\em states},\\
{$F$} & is a subset of $Q$, the set of {\em final states}, and\\
{$M$} & is a partial function from $\Sigma\times Q^*$ to
$\power(Q)$ (the powerset of $Q$) with finite domain,
the {\em transition function}.
\end{tabular}
\vspace{1ex}

The transition function $M$ associates sets of states with alphabet symbols.
It induces a function that associates sets of states with trees,
$\overline{M}:T_\Sigma\to\power(Q)$, such that:
\[
\begin{array}{rl}
\multicolumn{2}{l}{q\in\overline{M}(t) \defiff}\\
&t\mbox{ is a leaf labeled $\sigma$ and }q\in M(\sigma,\varepsilon), \tor\\
&t = \sigma(t_0,\ldots,t_n)\mbox{ and there is a sequence}\\
&\qquad\mbox{of states } q_0,\ldots,q_n \mbox{ such that }
q_i\in\overline{M}(t_i),\\
&\qquad \mbox{ for } 0\leq i\leq n, \tand
q\in M(\sigma,q_0\cdot\;\cdots\;\cdot q_n).
\end{array}
\]

An automaton $\sA=\tup{\Sigma,Q,M,F}$ accepts a tree $t\in T_\Sigma$
iff, by definition, $F\cap\overline{M}(t)$ is not empty.  The set of
trees accepted by an automaton $\sA$ is denoted $T(\sA)$.

A set of trees is {\em recognizable} iff, by definition, it is $T(\sA)$
for some automaton $\sA$.

\begin{lemma}\cite{thatch67}
Every local set is recognizable.  Every recognizable set is the
projection of some local set.
\end{lemma}
The projection is necessary because the automaton can distinguish
between nodes labeled with the same symbol while the CFG
cannot.  The set of trees (with bounded branching)
in which exactly one node is labeled $A$,
for instance, is recognizable but not local.
It is, however, the projection of a local set in which the labels of the
nodes that dominate the node labeled $A$ are distinguished from the labels
of those that don't.

As a corollary of this lemma, the {\em path set} of a
recognizable (or local) set,
i.e., the set of strings that label paths in the trees in
that set, is regular.

\begin{figure*}[tb]
\leavevmode
\centering
\begingroup\makeatletter
\def\x#1#2#3#4#5#6#7\relax{\def\x{#1#2#3#4#5#6}}%
\expandafter\x\fmtname xxxxxx\relax \def\y{splain}%
\ifx\x\y   
\gdef\SetFigFont#1#2#3{%
  \ifnum #1<17\tiny\else \ifnum #1<20\small\else
  \ifnum #1<24\normalsize\else \ifnum #1<29\large\else
  \ifnum #1<34\Large\else \ifnum #1<41\LARGE\else
     \huge\fi\fi\fi\fi\fi\fi
  \csname #3\endcsname}%
\else
\gdef\SetFigFont#1#2#3{\begingroup
  \count@#1\relax \ifnum 25<\count@\count@25\fi
  \def\x{\endgroup\@setsize\SetFigFont{#2pt}}%
  \expandafter\x
    \csname \romannumeral\the\count@ pt\expandafter\endcsname
    \csname @\romannumeral\the\count@ pt\endcsname
  \csname #3\endcsname}%
\fi
\endgroup
\begin{picture}(0,0)%
\includegraphics{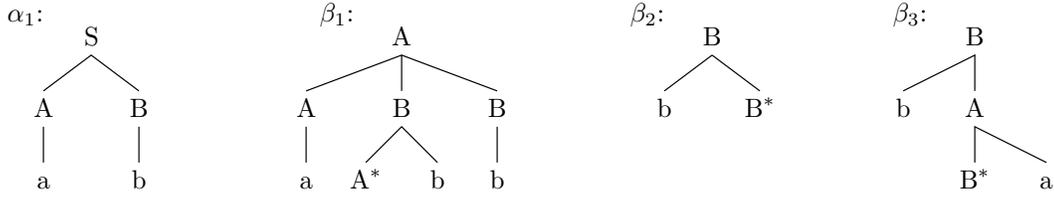}%
\end{picture}%
\setlength{\unitlength}{0.012500in}%
\begingroup\makeatletter
\def\x#1#2#3#4#5#6#7\relax{\def\x{#1#2#3#4#5#6}}%
\expandafter\x\fmtname xxxxxx\relax \def\y{splain}%
\ifx\x\y   
\gdef\SetFigFont#1#2#3{%
  \ifnum #1<17\tiny\else \ifnum #1<20\small\else
  \ifnum #1<24\normalsize\else \ifnum #1<29\large\else
  \ifnum #1<34\Large\else \ifnum #1<41\LARGE\else
     \huge\fi\fi\fi\fi\fi\fi
  \csname #3\endcsname}%
\else
\gdef\SetFigFont#1#2#3{\begingroup
  \count@#1\relax \ifnum 25<\count@\count@25\fi
  \def\x{\endgroup\@setsize\SetFigFont{#2pt}}%
  \expandafter\x
    \csname \romannumeral\the\count@ pt\expandafter\endcsname
    \csname @\romannumeral\the\count@ pt\endcsname
  \csname #3\endcsname}%
\fi
\endgroup
\begin{picture}(470,81)(-28,760)
\put( 40,820){\makebox(0,0)[b]{\smash{\SetFigFont{10}{12.0}{rm}S}}}
\put( 20,790){\makebox(0,0)[b]{\smash{\SetFigFont{10}{12.0}{rm}A}}}
\put( 20,760){\makebox(0,0)[b]{\smash{\SetFigFont{10}{12.0}{rm}a}}}
\put( 60,790){\makebox(0,0)[b]{\smash{\SetFigFont{10}{12.0}{rm}B}}}
\put( 60,760){\makebox(0,0)[b]{\smash{\SetFigFont{10}{12.0}{rm}b}}}
\put(170,820){\makebox(0,0)[b]{\smash{\SetFigFont{10}{12.0}{rm}A}}}
\put(170,790){\makebox(0,0)[b]{\smash{\SetFigFont{10}{12.0}{rm}B}}}
\put(130,760){\makebox(0,0)[b]{\smash{\SetFigFont{10}{12.0}{rm}a}}}
\put(210,760){\makebox(0,0)[b]{\smash{\SetFigFont{10}{12.0}{rm}b}}}
\put(130,790){\makebox(0,0)[b]{\smash{\SetFigFont{10}{12.0}{rm}A}}}
\put(210,790){\makebox(0,0)[b]{\smash{\SetFigFont{10}{12.0}{rm}B}}}
\put(185,760){\makebox(0,0)[b]{\smash{\SetFigFont{10}{12.0}{rm}b}}}
\put(300,820){\makebox(0,0)[b]{\smash{\SetFigFont{10}{12.0}{rm}B}}}
\put(280,790){\makebox(0,0)[b]{\smash{\SetFigFont{10}{12.0}{rm}b}}}
\put(410,820){\makebox(0,0)[b]{\smash{\SetFigFont{10}{12.0}{rm}B}}}
\put(380,790){\makebox(0,0)[b]{\smash{\SetFigFont{10}{12.0}{rm}b}}}
\put(410,790){\makebox(0,0)[b]{\smash{\SetFigFont{10}{12.0}{rm}A}}}
\put(440,760){\makebox(0,0)[b]{\smash{\SetFigFont{10}{12.0}{rm}a}}}
\put( 20,830){\makebox(0,0)[rb]{\smash{\SetFigFont{10}{12.0}{rm}$\alpha_1$:}}}
\put(150,830){\makebox(0,0)[rb]{\smash{\SetFigFont{10}{12.0}{rm}$\beta_1$:}}}
\put(280,830){\makebox(0,0)[rb]{\smash{\SetFigFont{10}{12.0}{rm}$\beta_2$:}}}
\put(390,830){\makebox(0,0)[rb]{\smash{\SetFigFont{10}{12.0}{rm}$\beta_3$:}}}
\put(155,760){\makebox(0,0)[b]{\smash{\SetFigFont{10}{12.0}{rm}$\f{A}^*$}}}
\put(320,790){\makebox(0,0)[b]{\smash{\SetFigFont{10}{12.0}{rm}$\f{B}^*$}}}
\put(410,760){\makebox(0,0)[b]{\smash{\SetFigFont{10}{12.0}{rm}$\f{B}^*$}}}
\end{picture}
\caption{Regular Adjunction}
\label{fig.regadj}
\end{figure*}
\section{TAGs in Regular Form}\label{sec.regular}
\subsection{Regular Adjunction}\label{sec.regadj}
The fact that the path sets of recognizable sets must be regular
provides our basic approach to defining a class of TAGs that generate
only recognizable sets.  We start with a restricted form of
adjunction that can generate only regular path sets and then look for a
class of TAGs that do not generate any trees that cannot be generated
with this restricted form of adjunction.

\begin{figure*}[tb]
\leavevmode
\centering
\begingroup\makeatletter
\def\x#1#2#3#4#5#6#7\relax{\def\x{#1#2#3#4#5#6}}%
\expandafter\x\fmtname xxxxxx\relax \def\y{splain}%
\ifx\x\y   
\gdef\SetFigFont#1#2#3{%
  \ifnum #1<17\tiny\else \ifnum #1<20\small\else
  \ifnum #1<24\normalsize\else \ifnum #1<29\large\else
  \ifnum #1<34\Large\else \ifnum #1<41\LARGE\else
     \huge\fi\fi\fi\fi\fi\fi
  \csname #3\endcsname}%
\else
\gdef\SetFigFont#1#2#3{\begingroup
  \count@#1\relax \ifnum 25<\count@\count@25\fi
  \def\x{\endgroup\@setsize\SetFigFont{#2pt}}%
  \expandafter\x
    \csname \romannumeral\the\count@ pt\expandafter\endcsname
    \csname @\romannumeral\the\count@ pt\endcsname
  \csname #3\endcsname}%
\fi
\endgroup
\begin{picture}(0,0)%
\includegraphics{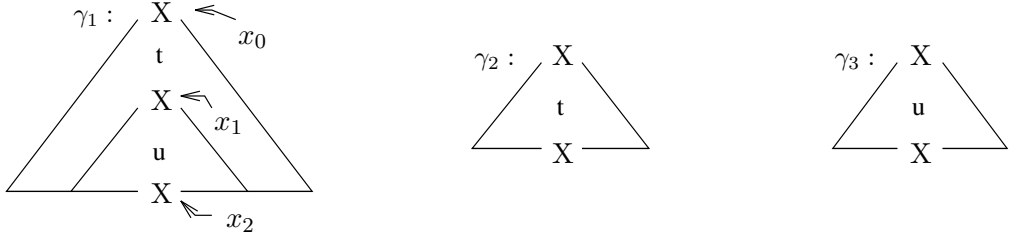}%
\end{picture}%
\setlength{\unitlength}{0.012500in}%
\begingroup\makeatletter
\def\x#1#2#3#4#5#6#7\relax{\def\x{#1#2#3#4#5#6}}%
\expandafter\x\fmtname xxxxxx\relax \def\y{splain}%
\ifx\x\y   
\gdef\SetFigFont#1#2#3{%
  \ifnum #1<17\tiny\else \ifnum #1<20\small\else
  \ifnum #1<24\normalsize\else \ifnum #1<29\large\else
  \ifnum #1<34\Large\else \ifnum #1<41\LARGE\else
     \huge\fi\fi\fi\fi\fi\fi
  \csname #3\endcsname}%
\else
\gdef\SetFigFont#1#2#3{\begingroup
  \count@#1\relax \ifnum 25<\count@\count@25\fi
  \def\x{\endgroup\@setsize\SetFigFont{#2pt}}%
  \expandafter\x
    \csname \romannumeral\the\count@ pt\expandafter\endcsname
    \csname @\romannumeral\the\count@ pt\endcsname
  \csname #3\endcsname}%
\fi
\endgroup
\begin{picture}(419,97)(0,740)
\put( 64,826){\makebox(0,0)[b]{\smash{\SetFigFont{11}{13.2}{rm}X}}}
\put( 27,826){\makebox(0,0)[lb]{\smash{\SetFigFont{10}{12.0}{rm}$\gamma_1:$}}}
\put( 64,789){\makebox(0,0)[b]{\smash{\SetFigFont{11}{13.2}{rm}X}}}
\put( 64,750){\makebox(0,0)[b]{\smash{\SetFigFont{11}{13.2}{rm}X}}}
\put( 96,817){\makebox(0,0)[lb]{\smash{\SetFigFont{11}{13.2}{rm}$x_0$}}}
\put( 86,781){\makebox(0,0)[lb]{\smash{\SetFigFont{11}{13.2}{rm}$x_1$}}}
\put( 91,740){\makebox(0,0)[lb]{\smash{\SetFigFont{11}{13.2}{rm}$x_2$}}}
\put(382,808){\makebox(0,0)[b]{\smash{\SetFigFont{11}{13.2}{rm}X}}}
\put(382,767){\makebox(0,0)[b]{\smash{\SetFigFont{11}{13.2}{rm}X}}}
\put(346,808){\makebox(0,0)[lb]{\smash{\SetFigFont{10}{12.0}{rm}$\gamma_3:$}}}
\put(195,808){\makebox(0,0)[lb]{\smash{\SetFigFont{10}{12.0}{rm}$\gamma_2:$}}}
\put(232,808){\makebox(0,0)[b]{\smash{\SetFigFont{11}{13.2}{rm}X}}}
\put(232,767){\makebox(0,0)[b]{\smash{\SetFigFont{11}{13.2}{rm}X}}}
\end{picture}
\caption{Regular Form}
\label{fig.regform}
\end{figure*}

\begin{definition}
{\em\bf Regular adjunction} is ordinary 
adjunction restricted to the following cases:
\begin{itemize}
\item any auxiliary tree may be adjoined into any initial tree or at any
node that is not on the spine of an auxiliary tree,
\item any {\em proper}
auxiliary tree may be adjoined into any auxiliary tree at the
root or foot of that tree,
\item any auxiliary tree $\gamma_1$ may be adjoined at any node 
along the spine of
any auxiliary tree $\gamma_2$ provided that no instance of $\gamma_2$
can be adjoined at any node along the spine of $\gamma_1$.
\end{itemize}
\end{definition}
In figure~\ref{fig.regadj}, for example, this rules out adjunction of
$\beta_1$ into the spine of $\beta_3$, or {\em vice versa}, either
directly or indirectly (by adjunction of $\beta_3$, say,
into $\beta_2$ and then
adjunction of the resulting auxiliary tree into $\beta_1$.)  Note that,
in the case of TAGs with no improper elementary auxiliary trees, the
requirement that only proper auxiliary trees may be adjoined at the root
or foot is not actually a restriction.  This is because the only way to
derive an improper auxiliary tree in such a TAG without violating the
other restrictions on regular adjunction is by adjunction at the root or
foot.  Any sequence of such adjunctions can always be re-ordered in a
way which meets the requirement.

We denote the set of completed initial trees derivable by regular
adjunction in $G$ as $T_R(G)$.  Similarly, we denote the set of all
trees that are derivable by regular adjunction
in $G$ as $T'_R(G)$.  As intended, we can show that $T_R(G)$ is always a
recognizable set.  We are looking, then, for a class of TAGs for which
$T(G)=T_R(G)$ for every $G$ in the class.  Clearly, this will be the
case if $T'(G)=T'_R(G)$ for every such $G$.

\begin{proposition}\label{prop.reg.implies.recog}
If $G$ is a TAG and $T'(G)=T'_R(G)$. Then $T(G)$ is a recognizable set.
\end{proposition}
\begin{proofof}{Sketch}
This follows from the fact that in regular adjunction, if one treats
adjunction at the root or foot as substitution,
there is a fixed bound, dependent only on $G$, on the depth to which
auxiliary trees can be nested.  Thus the nesting of the auxiliary trees
can be tracked by a fixed depth stack.  Such a stack can be encoded in a
finite set of states.  It's reasonably easy to see, then, how $G$ can be
compiled into a bottom-up finite state tree automaton.
\end{proofof}

Since regular adjunction generates only recognizable sets, and thus
(modulo projection) local sets, and since CFGs
can be parsed in cubic time, one would hope that TAGs that employ only
regular adjunction can be parsed in cubic time as well.  In fact, such
is the case.

\begin{proposition}
If $G$ is a TAG for which $T(G)=T_R(G)$ then there is a algorithm that
recognizes strings in $L(G)$ in time proportional to the cube of the
length of the string.\footnote{This result was suggested by K.
Vijay-Shanker.} 
\end{proposition}
\begin{proofof}{Sketch}
This, again, follows from the fact that the depth of nesting of
auxiliary trees is bounded in regular adjunction.  A
CKY-style style parsing algorithm for TAGs (the one given
in~\newcite{VijWei93}, for example) can be modified to
work with a two-dimensional array, storing in each slot $[i,j]$
a set of structures that encode a node in an elementary tree that can
occur at the root of a subtree spanning the input from position $i$
through $j$ in some tree derivable in $G$, along with a stack recording
the nesting of elementary auxiliary trees around that node in the
derivation of that tree.
Since the stacks are bounded the amount of data
stored in each node is independent of the input length and the
algorithm executes in time proportional to the cube of the length of the
input. 
\end{proofof}

\subsection{Regular Form}\label{sec.regform}
We are interested in classes of TAGs for which $T'(G)=T'_R(G)$.  One
such class is the TAGs in {\em regular form}.
\begin{definition}
A TAG is in {\em\bf regular form} iff whenever a {\em completed}
auxiliary tree of the form $\gamma_1$ in Figure~\ref{fig.regform} is
derivable, where
$x_0\not= x_1\not= x_2$ and
no node labeled $\f{X}$ occurs properly between
$x_0$ and $x_1$,
then trees of the form $\gamma_2$ and $\gamma_3$ are
derivable as well.
\end{definition}
Effectively, this is a closure condition on the
elementary trees of the grammar.  Note that it
immediately implies that every
improper elementary auxiliary tree in a regular form TAG is redundant.
It is also easy to see, by induction on the
number of occurrences of $X$ along the spine, that any auxiliary tree
$\gamma$ for $X$ that is derivable in $G$
can be decomposed into the concatenation of a
sequence of proper auxiliary trees for $X$ each of which is
derivable in $G$.  We will refer to the proper auxiliary trees in this
sequence as the {\em proper segments} of $\gamma$.

\begin{lemma}\label{lem.reg.form.implies.reg.adj}
Suppose $G$ is a TAG in regular form.  Then $T'(G)=T'_R(G)$
\end{lemma}
\begin{proof}
Suppose $\gamma$ is any non-elementary
auxiliary tree derivable by unrestricted
adjunction in $G$ and that any smaller tree derivable in $G$ is
derivable by regular adjunction in $G$.  If $\gamma$ is proper,
then it is clearly derivable from two strictly smaller trees by
regular adjunction, each of which, by the induction hypothesis, is in
$T'_R(G)$.  If $\gamma$ is improper, then it has the form of
$\gamma_1$ in Figure~\ref{fig.regform} and it is derivable by regular
adjunction of $\gamma_2$ at the root of $\gamma_3$.  Since both of these
are derivable and strictly smaller than $\gamma$ they are in $T'_R(G)$.
It follows that $\gamma$ is in $T'_R(G)$ as well.
\end{proof}

\begin{lemma}\label{lem.reg.adj.implies.reg.form}
Suppose $G$ is a TAG with no improper elementary trees
and $T'(G)=T'_R(G)$.  Then $G$ is in regular form.
\end{lemma}
\begin{proof}
Suppose some $\gamma$ with the form of $\gamma_1$ in
Figure~\ref{fig.regform} is derivable in $G$
and that for all trees $\gamma'$ that are smaller than $\gamma$ every
proper segment of $\gamma'$ is derivable in $G$. 
By assumption $\gamma$ is not elementary
since it is improper.  Thus, by hypothesis, $\gamma$ is derivable by
regular adjunction of some $\gamma''$ into some $\gamma'$ both of which
are derivable in $G$.

Suppose $\gamma''$ adjoins into the spine of $\gamma'$ and that
a node labeled $X$ occurs along the spine
of $\gamma''$.  Then, by the definition of regular adjunction, $\gamma''$
must be adjoined at either the root or foot of $\gamma'$.  Thus
both $\gamma'$ and $\gamma''$ consist of sequences of consecutive proper
segments of $\gamma$ with $\gamma''$ including $t$ and the initial
(possibly empty) portion
of $u$ and $\gamma'$ including the remainder of $u$ or vice
versa.  In either case, by the induction hypothesis, every proper
segment of both $\gamma'$ and $\gamma''$, and thus every proper segment
of $\gamma$ is derivable in $G$.  Then trees of the form $\gamma_2$ and
$\gamma_3$ are derivable from these proper segments.

Suppose, on the other hand,
that $\gamma''$ does not adjoin along the spine of $\gamma'$ or that
no node labeled $X$ occurs along the spine of $\gamma''$.  Note
that $\gamma''$ must occur entirely within a proper segment of
$\gamma$.  Then
$\gamma'$ is a tree with the form of $\gamma_1$ that is smaller than
$\gamma$.  From the induction hypothesis every proper segment of
$\gamma'$ is derivable in $G$.
It follows then that every proper segment of $\gamma$ is derivable in
$G$, either because it is a proper segment of $\gamma'$ or because it is
derivable by adjunction of $\gamma''$ into a proper segment of
$\gamma'$.  Again, trees of the form $\gamma_2$ and $\gamma_3$ are
derivable from these proper segments.
\end{proof}

\subsection{Regular Form and Local Sets}\label{sec.reg.local}
The class of TAGs in regular form is related to the local sets in much
the same way that the class of regular grammars is related to regular
languages.  Every TAG in regular form generates a recognizable set.
This follows from Lemma~\ref{lem.reg.form.implies.reg.adj} and
Proposition~\ref{prop.reg.implies.recog}.
Thus, modulo projection, every TAG in regular form generates a local
set.  Conversely, the next proposition establishes that
every local set 
can be generated by a TAG in regular
form.  Thus regular form provides a normal form for TAGs that
generate local sets.  It is not the case, however, that all TAGs that
generate local sets are in regular form.

\begin{proposition}
For every CFG $G$ there is a TAG $G'$ in regular form such that the set
of derivation trees for $G$ is exactly $T(G')$.
\end{proposition}
\begin{proof}
This is nearly immediate, since every CFG is equivalent to a
Tree Substitution Grammar (in which all trees are of depth one) and
every Tree Substitution Grammar is, in the definition we use here, a TAG
with no elementary auxiliary trees.  It follows that this TAG can derive
no auxiliary trees at all, and is thus vacuously in regular form.
\end{proof}
This proof is hardly satisfying, depending as it does on the fact that
TAGs, as we define them, can employ substitution.  The next proposition
yields, as a corollary, the more substantial result that every CFG is
strongly equivalent to a TAG in regular form in which substitution
plays no role.

\begin{proposition}
The class of TAGs in regular form can lexicalize CFGs.
\end{proposition}
\begin{proof}
This follows directly from the equivalent lemma in~\newcite{SchWat93a}.
The construction given there builds a
{\em left-corner derivation graph} (LCG). 
Vertices in this graph are the terminals and non-terminals of $G$. 
Edges correspond to the productions of $G$ in the following way: there
is an edge from $X$ to $Y$ labeled $X\to Y\alpha$ iff $X\to Y\alpha$ is
a production in $G$.
Paths through this graph that end on a terminal
characterize the left-corner derivations in $G$.  The construction
proceeds by building a set of elementary initial trees corresponding to
the simple (acyclic) paths through the LCG that end on terminals. 
These capture the non-recursive left-corner derivations in $G$.  The set
of auxiliary trees is built in two steps.  First, an auxiliary tree is
constructed for every simple cycle in the graph.  This gives a set
of auxiliary trees that is sufficient, with the initial trees, to derive
every tree generated by the CFG.  This set of auxiliary trees, however,
may include some which are not lexicalized, that is, in which
every frontier node other than the foot is
marked for substitution.  These can be lexicalized by
substituting every corresponding elementary initial tree at
one of those frontier nodes.
Call the LCFG constructed for $G$ by this method $G'$.
For our purposes, the important point of the construction is that every
simple cycle in the LCG is represented by an elementary auxiliary tree. 
Since the spines of auxiliary trees derivable in $G'$ correspond to
cycles in the LCG, every proper segment of an auxiliary
tree derivable in $G'$ is a simple cycle in the LCG.  Thus every such
proper segment is derivable in $G'$ and $G'$ is in regular form.
\end{proof}

The use of a graph which captures left-corner derivations as the
foundation of this construction guarantees that the auxiliary trees it
builds will be left-recursive (will have the foot as the left-most
leaf.)  It is a requirement of LCFGs that all auxiliary trees be either
left- or right-recursive.  Thus, while other derivation strategies may
be employed in constructing the graph, these must always expand either
the left- or right-most child at each step.  All that is required for
the construction to produce a TAG in regular form, though, is that every
simple cycle in the graph be realized in an elementary tree.
The resulting grammar will be in regular form no
matter what (complete) derivation strategy is captured in the graph.  In
particular, this admits the possibility of generating an LTAG in which
the anchor of each elementary tree is some linguistically motivated
``head''. 

\begin{corollary}
For every CFG $G$ there is a TAG $G'$ in regular form {\em in which no
node is marked for substitution}, such that the set
of derivation trees for $G$ is exactly $T(G')$.
\end{corollary}
This follows from the fact that the step used to lexicalize the
elementary auxiliary trees in Schabes and Waters's construction can be
applied to every node (in both initial and auxiliary trees) which is
marked for substitution.  Paradoxically, to establish the corollary it
is not necessary for every elementary tree to be lexicalized.  
In Schabes and Waters's lemma $G$ is required to be finitely ambiguous
and to not generate the empty string.  These restrictions are only
necessary if $G'$ is to be lexicalized.  Here we can accept TAGs which
include elementary trees in which the only leaf is the foot node or
which yield only the empty string.  Thus the corollary applies to all
CFGs without restriction.

\subsection{Regular Form is Decidable}\label{sec.decide}
\begin{figure*}[tb]
\leavevmode
\hspace*{30pt}
\begingroup\makeatletter
\def\x#1#2#3#4#5#6#7\relax{\def\x{#1#2#3#4#5#6}}%
\expandafter\x\fmtname xxxxxx\relax \def\y{splain}%
\ifx\x\y   
\gdef\SetFigFont#1#2#3{%
  \ifnum #1<17\tiny\else \ifnum #1<20\small\else
  \ifnum #1<24\normalsize\else \ifnum #1<29\large\else
  \ifnum #1<34\Large\else \ifnum #1<41\LARGE\else
     \huge\fi\fi\fi\fi\fi\fi
  \csname #3\endcsname}%
\else
\gdef\SetFigFont#1#2#3{\begingroup
  \count@#1\relax \ifnum 25<\count@\count@25\fi
  \def\x{\endgroup\@setsize\SetFigFont{#2pt}}%
  \expandafter\x
    \csname \romannumeral\the\count@ pt\expandafter\endcsname
    \csname @\romannumeral\the\count@ pt\endcsname
  \csname #3\endcsname}%
\fi
\endgroup
\begin{picture}(0,0)%
\includegraphics{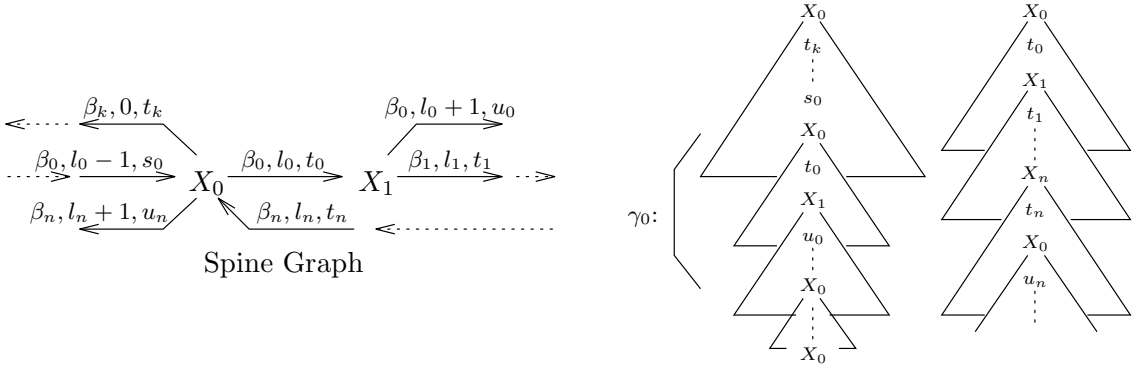}%
\end{picture}%
\setlength{\unitlength}{0.012500in}%
\begingroup\makeatletter
\def\x#1#2#3#4#5#6#7\relax{\def\x{#1#2#3#4#5#6}}%
\expandafter\x\fmtname xxxxxx\relax \def\y{splain}%
\ifx\x\y   
\gdef\SetFigFont#1#2#3{%
  \ifnum #1<17\tiny\else \ifnum #1<20\small\else
  \ifnum #1<24\normalsize\else \ifnum #1<29\large\else
  \ifnum #1<34\Large\else \ifnum #1<41\LARGE\else
     \huge\fi\fi\fi\fi\fi\fi
  \csname #3\endcsname}%
\else
\gdef\SetFigFont#1#2#3{\begingroup
  \count@#1\relax \ifnum 25<\count@\count@25\fi
  \def\x{\endgroup\@setsize\SetFigFont{#2pt}}%
  \expandafter\x
    \csname \romannumeral\the\count@ pt\expandafter\endcsname
    \csname @\romannumeral\the\count@ pt\endcsname
  \csname #3\endcsname}%
\fi
\endgroup
\begin{picture}(471,152)(5,670)
\put( 89,741){\makebox(0,0)[b]{\smash{\SetFigFont{11}{13.2}{rm}$X_0$}}}
\put(160,741){\makebox(0,0)[b]{\smash{\SetFigFont{11}{13.2}{rm}$X_1$}}}
\put( 54,773){\makebox(0,0)[b]{\smash{\SetFigFont{9}{10.8}{rm}$\beta_k,0,t_k$}}}
\put(191,751){\makebox(0,0)[b]{\smash{\SetFigFont{9}{10.8}{rm}$\beta_1,l_1,t_1$}}}
\put( 44,729){\makebox(0,0)[b]{\smash{\SetFigFont{9}{10.8}{rm}$\beta_n,l_n+1,u_n$}}}
\put(129,729){\makebox(0,0)[b]{\smash{\SetFigFont{9}{10.8}{rm}$\beta_n,l_n,t_n$}}}
\put(121,707){\makebox(0,0)[b]{\smash{\SetFigFont{11}{13.2}{rm}Spine Graph}}}
\put( 44,751){\makebox(0,0)[b]{\smash{\SetFigFont{9}{10.8}{rm}$\beta_0,l_0-1,s_0$}}}
\put(121,751){\makebox(0,0)[b]{\smash{\SetFigFont{9}{10.8}{rm}$\beta_0,l_0,t_0$}}}
\put(191,773){\makebox(0,0)[b]{\smash{\SetFigFont{9}{10.8}{rm}$\beta_0,l_0+1,u_0$}}}
\put(436,814){\makebox(0,0)[b]{\smash{\SetFigFont{7}{8.4}{rm}$X_0$}}}
\put(436,785){\makebox(0,0)[b]{\smash{\SetFigFont{7}{8.4}{rm}$X_1$}}}
\put(436,771){\makebox(0,0)[b]{\smash{\SetFigFont{7}{8.4}{rm}$t_1$}}}
\put(436,746){\makebox(0,0)[b]{\smash{\SetFigFont{7}{8.4}{rm}$X_n$}}}
\put(436,731){\makebox(0,0)[b]{\smash{\SetFigFont{7}{8.4}{rm}$t_n$}}}
\put(436,717){\makebox(0,0)[b]{\smash{\SetFigFont{7}{8.4}{rm}$X_0$}}}
\put(436,702){\makebox(0,0)[b]{\smash{\SetFigFont{7}{8.4}{rm}$u_n$}}}
\put(436,800){\makebox(0,0)[b]{\smash{\SetFigFont{7}{8.4}{rm}$t_0$}}}
\put(343,699){\makebox(0,0)[b]{\smash{\SetFigFont{7}{8.4}{rm}$X_0$}}}
\put(343,670){\makebox(0,0)[b]{\smash{\SetFigFont{7}{8.4}{rm}$X_0$}}}
\put(343,814){\makebox(0,0)[b]{\smash{\SetFigFont{7}{8.4}{rm}$X_0$}}}
\put(343,800){\makebox(0,0)[b]{\smash{\SetFigFont{7}{8.4}{rm}$t_k$}}}
\put(343,764){\makebox(0,0)[b]{\smash{\SetFigFont{7}{8.4}{rm}$X_0$}}}
\put(343,735){\makebox(0,0)[b]{\smash{\SetFigFont{7}{8.4}{rm}$X_1$}}}
\put(343,778){\makebox(0,0)[b]{\smash{\SetFigFont{7}{8.4}{rm}$s_0$}}}
\put(343,749){\makebox(0,0)[b]{\smash{\SetFigFont{7}{8.4}{rm}$t_0$}}}
\put(343,720){\makebox(0,0)[b]{\smash{\SetFigFont{7}{8.4}{rm}$u_0$}}}
\put(278,728){\makebox(0,0)[rb]{\smash{\SetFigFont{9}{10.8}{rm}$\gamma_0$:}}}
\end{picture}
\caption{Regular Form is Decidable}
\label{fig.segments}
\end{figure*}
We have established that regular form gives a class of TAGs that is
strongly equivalent to CFGs (modulo projection),
and that LTAGs in this class lexicalize CFGs.  In this section we 
provide an effective procedure for deciding if a given TAG is in
regular form.
The procedure is based on a graph
that is not unlike the LCG of the construction of Schabes and Waters.

If $G$ is a TAG, the {\em Spine Graph} of $G$ is a directed multi-graph
on a set of vertices, one for each non-terminal in $G$.  If $\beta_i$ is
an elementary auxiliary tree in $G$ and the spine of
$\beta_i$ is labeled with the sequence of
non-terminals $\tup{X_0,X_1,\ldots,X_n}$ (where $X_0=X_n$ and the
remaining $X_j$ are not necessarily distinct),
then there is an edge in the graph from each $X_j$ to $X_{j+1}$
labeled $\tup{\beta_i,j,t_{i,j}}$, where $t_{i,j}$ is that portion of
$\beta_i$ that is dominated by $X_j$ but not properly dominated by
$X_{j+1}$.  There are no other edges in the graph
except those corresponding to the elementary auxiliary trees of $G$ in
this way.

The intent is for the spine graph of $G$ to characterize the set of
auxiliary trees derivable in $G$ by adjunction along the spine.
Clearly, any vertex
that is labeled with a non-terminal for which there
is no corresponding auxiliary tree plays no active role in these
derivations and can be replaced, along with the
pairs of edges incident on it, by single edges.  Without loss of
generality, then, we assume
spine graphs of this reduced form.  Thus every vertex has at least one
edge labeled with a 0 in its second component incident from it.

A {\em well-formed-cycle} (wfc) in this graph is a (non-empty)
path traced by
the following non-deterministic automaton:
\begin{itemize}
\item The automaton consists of a single push-down stack.  Stack
contents are labels of edges in the graph.
\item The automaton starts on any vertex of the graph with an empty stack.
\item At each step, the automaton can move as follows:
\begin{itemize}
\item If there is an edge incident from the current vertex labeled
$\tup{\beta_i,0,t_{i,0}}$
the automaton can push that label onto the stack and
move to the vertex at the far end of that edge.
\item If the top of stack contains $\tup{\beta_i,j,t_{i,j}}$
and there is an
edge incident from the current vertex labeled $\tup{\beta_i,j+1,t_{i,j+1}}$
the
automaton may pop the top of stack, push $\tup{\beta_i,j+1,t_{i,j+1}}$
and move to the vertex at the end of that edge.
\item If the top of stack contains $\tup{\beta_i,j,t_{i,j}}$ but there is no
edge incident from the current vertex labeled $\tup{\beta_i,j+1,t_{i,j+1}}$
then the automaton may pop the top of stack and remain at the same vertex.
\end{itemize}
\item The automaton may halt if its stack is empty.
\item A path through the graph is traced by the automaton if it starts
at the first vertex in the path and halts at the last vertex in the path
visiting each of the vertices in the path in order.
\end{itemize}

Each wfc in a spine graph corresponds to the auxiliary tree built
by concatenating
the third components of the labels on the edges in the cycle in order. 
Then every wfc in the spine graph of $G$ corresponds to an auxiliary
tree that is derivable in $G$ by adjunction along the spine only. 
Conversely, every such auxiliary tree corresponds to some wfc in the
spine graph.

A {\em simple cycle} in the spine graph, by definition, is any minimal
cycle 
in the graph that ignores the labels of the edges but not their
direction.  Simple cycles correspond to auxiliary trees in the same way
that wfcs do.  Say that two cycles in the graph are equivalent iff they
correspond to the same auxiliary tree.
The simple cycles in the spine graph for $G$
correspond to the minimal set of elementary auxiliary trees in any
presentation of $G$ that is closed under the regular form condition in
the following way.

\begin{lemma}\label{lem.spgraph}
A TAG $G$ is in regular form iff every
simple cycle in its spine graph is equivalent to a wfc in that
graph. 
\end{lemma}
\begin{proof}\
\newline
(If every simple cycle is equivalent to a wfc then $G$ is in regular form.)

Suppose every simple cycle in the spine graph of $G$ is equivalent to
a wfc and
some tree of the form $\gamma_1$ in Figure~\ref{fig.regform} is
derivable in $G$.  Wlog, assume the tree is derivable by adjunction
along the spine only.
Then there is a wfc in the spine graph of $G$ corresponding to that tree
that is of the
form $\tup{X_0,\ldots,X_k,\ldots,X_n}$ where $X_0=X_k=X_n$, $0\not=k\not=n$,
and
$X_i\not=X_0$ for all $0<i<k$.  Thus $\tup{X_0,\ldots,X_k}$ is a simple
cycle in the spine graph.  Further, $\tup{X_k,\ldots,X_n}$ is a
sequence of one or more such simple cycles.  It follows
that both $\tup{X_0,\ldots,X_k}$ and $\tup{X_k,\ldots,X_n}$ are wfc in
the spine graph and thus both $\gamma_2$ and $\gamma_3$ are derivable in
$G$.

\noindent
(If $G$ is in regular form then every simple cycle corresponds to a wfc.)

Assume, wlog, the spine graph of $G$ is connected.  (If it is not we
can treat $G$ as a union of grammars.)  Since the spine graph is a union
of wfcs it has an Eulerian wfc (in the usual sense of
Eulerian).  Further, since every vertex is the initial vertex of some
wfc, every vertex is the initial vertex of some Eulerian wfc.

Suppose there is some simple cycle
\[
\begin{array}{l}
X_0\,\tup{\beta_0,l_0,t_0}\,
X_1\,\tup{\beta_1,l_1,t_1}\,\cdots\\
\quad\cdots\, X_n\,\tup{\beta_n,l_n,t_n}\,X_0
\end{array}
\]
where the $X_j$ are the vertices and the tuples are the labels on the
edges of the cycle.  Then there is a wfc starting at $X_0$ that includes
the edge $\tup{\beta_0,l_0,t_0}$, although not necessarily initially.
In particular the
Eulerian wfc starting at $X_0$ is such a wfc.
This corresponds to a derivable
auxiliary tree that includes a proper segment beginning with
$t_0$.  Since $G$ is in regular form, that proper segment is a derivable
auxiliary tree.  Call this $\gamma_0$ (see Figure~\ref{fig.segments}.)
The spine of that tree is labeled $X_0,X_1,\ldots,X_0$,
where anything (other than $X_0$) can occur in the ellipses.

The same cycle can be rotated to get a simple cycle starting at each of
the $X_j$.  Thus for each $X_j$ there is a derivable auxiliary tree
starting with $t_j$.  Call it $\gamma_j$.  By a sequence of
adjunctions of each $\gamma_j$ at the second node on the spine of
$\gamma_{j-1}$ an auxiliary tree for $X_0$ is derivable in which the
first proper segment is the concatenation of
\[
t_0,t_1,\ldots,t_n.
\]
Again, by the fact that $G$ is in regular form, this proper
segment is derivable in $G$.  Hence there is a wfc in the spine graph
corresponding to this tree.
\end{proof}

\begin{proposition}\label{prop.rfdecidable}
For any TAG $G$ the question of whether $G$ is in regular form is
decidable.  Further, there is an effective procedure that, given any
TAG, will extend it to a TAG that is in regular form. 
\end{proposition}
\begin{proof}
Given a TAG $G$ we construct its
spine graph.  Since the TAG
is finite, the graph is as well.  The TAG is in regular form iff every
simple cycle is equivalent to 
a wfc.  This is clearly decidable.  Further, the set of
elementary trees corresponding to simple cycles that are not
equivalent to wfcs is
effectively constructible.
Adding that set to the original TAG extends it to regular form. 
\end{proof}
Of course the set of trees generated by the extended TAG may well be a
proper superset of the set generated by the original TAG.

\section{Discussion}\label{sec.disc}
The LCFGs of Schabes and Waters
employ a restricted form of adjunction and a
highly restricted form of elementary auxiliary tree.
The auxiliary
trees
of LCFGs
can only occur in left- or right-recursive form, that is,
with the foot as either the left- or right-most node on the  frontier of
the tree.  Thus the structures that can be captured in these
trees are restricted by the mechanism itself, and Schabes and Waters
(in~\shortcite{SchWat93a})
cite two situations where an existing LTAG
grammar for English~\cite{AbEtAl90} fails to meet this restriction.
But while it is sufficient to assure that the
language generated is context-free and cubic-time parsable,
this restriction is stronger than necessary.

TAGs in regular form, in contrast, are ordinary TAGs utilizing ordinary
adjunction.
While it is developed from the notion of regular
adjunction, regular form is just a closure condition on the elementary
trees of the grammar.
Although
that closure
condition assures that all
improper elementary auxiliary trees are redundant, the form of the
elementary trees themselves is unrestricted.  Thus the structures they
capture  can be driven
primarily by
linguistic considerations.
As we noted earlier, the restrictions on the form of the trees in an
LCFG significantly constrain the way in which CFGs can be lexicalized
using Schabes and Waters's construction.  These constraints are
eliminated if we require only that the result be in regular form and the
lexicalization can then be structured largely on linguistic principles.

On the other hand, regular form is a
property of the grammar as a whole, while the restrictions of LCFG are
restrictions on individual trees (and the manner in which they are
combined.)  Consequently, it is immediately obvious if a grammar meets
the requirements of LCFG, while it is less apparent if it is in regular
form.
In the case of the LTAG grammar for English, neither of the situations
noted by Schabes and Waters violate regular form themselves.  As
regular form is decidable, it is reasonable to
ask whether the grammar as a whole is in regular form.
A positive result would identify
the large fragment of English covered by this grammar as strongly
context-free and cubic-time parsable.  A negative result is likely to
give insight into those structures covered by the grammar that require
context-sensitivity. 

One might approach defining a context-free language within the TAG
formalism by developing a grammar with the intent that all trees derivable
in the grammar be derivable by regular adjunction.  This condition can
then be verified by the algorithm of previous section.  In the
case that the grammar is not in regular form, the algorithm proposes a
set of additional auxiliary trees that will establish that form.  In
essence, this is a prediction about the strings that would occur in a
context-free language extending the language encoded by the original
grammar.  It is then a linguistic issue whether these additional strings
are consistent with the intent of the grammar.

If a grammar is not in regular form, it is not necessarily the case
that it does not generate a recognizable set.  The main unresolved issue
in this work is whether it is possible to characterize the class of TAGs
that generate local sets more completely.  It is easy to show, for TAGs that
employ adjoining constraints,  that this is not
possible.  This is a consequence of the fact that one can construct, for any
CFG, a TAG in which the path language is the image, under a bijective
homomorphism,
of the string language generated by that CFG.  Since it is undecidable if
an arbitrary CFG generates a regular string language, and since the path
language of every recognizable set is regular, it is undecidable if an
arbitrary TAG (employing adjoining constraints) generates a recognizable
set.  This ability to capture CFLs in the string language, however,
seems to depend crucially on the nature of the adjoining constraints. 
It does not appear to extend to pure TAGs, or even TAGs in which the
adjoining constraints are implemented as monotonically growing sets of
simple features.  In the case of TAGs with these limited adjoining
constraints, then, the questions of whether there is a class of TAGs which
includes all and only those which generate recognizable sets, or if
there is an effective procedure for reducing any such TAG which generates
a recognizable set to one in regular form, are open.

\newpage


\begin{thebibliography}{}

\bibitem[\protect\citename{Abeill{\'e} \bgroup et al.\egroup }1990]{AbEtAl90}
Anne Abeill{\'e}, Kathleen~M. Bishop, Sharon Cote, and Yves Schabes.
\newblock 1990.
\newblock A lexicalized tree adjoining grammar for {E}nglish.
\newblock Technical Report MS-CIS-90-24, Department of Computer and Information
  Science, University of Pennsylvania.

\bibitem[\protect\citename{G{\'e}cseg and Steinby}1984]{GecSte84}
Ferenc G{\'e}cseg and Magnus Steinby.
\newblock 1984.
\newblock {\em Tree Automata}.
\newblock Akad{\'e}miai Kiad{\'o}, Budapest.

\bibitem[\protect\citename{Joshi and Schabes}1992]{JosSch92}
Aravind~K. Joshi and Yves Schabes.
\newblock 1992.
\newblock Tree-adjoining grammars and lexicalized grammars.
\newblock In M.~Nivat and A.~Podelski, editors, {\em Tree Automata and
  Languages}, pages 409--431. Elsevier Science Publishers B.V.

\bibitem[\protect\citename{Schabes and Joshi}1991]{SchJos91}
Yves Schabes and Aravind~K. Joshi.
\newblock 1991.
\newblock Parsing with lexicalized tree adjoining grammar.
\newblock In Masaru Tomita, editor, {\em Current Issues in Parsing Technology},
  chapter~3, pages 25--47. Kluwer Academic Publishers.

\bibitem[\protect\citename{Schabes and Waters}1993a]{SchWat93a}
Yves Schabes and Richard~C. Waters.
\newblock 1993a.
\newblock Lexicalized context-free grammars.
\newblock In {\em 31st Annual Meeting of the Association for Computational
  Linguistics (ACL'93)}, pages 121--129, Columbus, OH. Association for
  Computational Linguistics.

\bibitem[\protect\citename{Schabes and Waters}1993b]{SchWat93b}
Yves Schabes and Richard~C. Waters.
\newblock 1993b.
\newblock Lexicalized context-free grammar: {A} cubic-time parsable,
  lexicalized normal form for context-free grammar that preserves tree
  structure.
\newblock Technical Report 93-04, Mitsubishi Electric Research Laboratories
  Cambridge Research Center, Cambridge, MA, June.

\bibitem[\protect\citename{Schabes \bgroup et al.\egroup }1988]{ScAbJo88}
Yves Schabes, Anne Abeill{\'e}, and Aravind~K. Joshi.
\newblock 1988.
\newblock Parsing strategies with {`}lexicalized{'} grammars: Application to
  tree adjoining grammars.
\newblock In {\em Proceedings of the 12th International Conference on
  Computational Linguistics ({COLING'88})}, Budapest, Hungary. Association for
  Computational Linguistics.

\bibitem[\protect\citename{Schabes}1990]{schabe90}
Yves Schabes.
\newblock 1990.
\newblock {\em Mathematical and Computational Aspects of Lexicalized Grammars}.
\newblock {Ph.D.} thesis, Department of Computer and Information Science,
  University of Pennsylvania.

\bibitem[\protect\citename{Thatcher}1967]{thatch67}
J.~W. Thatcher.
\newblock 1967.
\newblock Characterizing derivation trees of context-free grammars through a
  generalization of finite automata theory.
\newblock {\em Journal of Computer and System Sciences}, 1:317--322.

\bibitem[\protect\citename{Vijay-Shanker and Weir}1993]{VijWei93}
K.~Vijay-Shanker and David Weir.
\newblock 1993.
\newblock Parsing some constrained grammar formalisms.
\newblock {\em Computational Linguistics}, 19(4):591--636.

\end{thebibliography}

\end{document}